\documentclass[runningheads]{llncs}

\usepackage{eccv}

\usepackage{eccvabbrv}

\usepackage{graphicx}
\usepackage{booktabs}

\usepackage[accsupp]{axessibility}  % Improves PDF readability for those with disabilities.

\usepackage{url}
\usepackage{amsmath, amssymb}
\usepackage{balance}
\usepackage{graphicx}
\usepackage{amsmath,graphicx}
\usepackage{booktabs} 
\usepackage{makecell} 
\usepackage{CJKutf8}
\usepackage{cite}
\usepackage{multirow}
\usepackage{amssymb}
\usepackage{booktabs}
\usepackage{arydshln}
\usepackage{hyperref}
\usepackage{indentfirst}
\usepackage{marvosym}
\usepackage{etoolbox}
\usepackage{threeparttable}
\usepackage{algorithm}
\usepackage{algorithmic}
\usepackage{adjustbox}
\usepackage{array}
\usepackage{color}

\usepackage{graphicx}
\usepackage{subcaption}
\begin{document}

% ---------------------------------------------------------------
% TODO REVIEW: Replace with your title
\title{FRRffusion: Unveiling Authenticity with Diffusion-Based Face Retouching Reversal} 

% TODO REVIEW: If the paper title is too long for the running head, you can set
% an abbreviated paper title here. If not, comment out.
\titlerunning{FRRffusion}

% TODO FINAL: Replace with your author list. 
% Include the authors' OCRID for the camera-ready version, if at all possible.
\author{Fengchuang Xing \and
Xiaowen Shi \and Yuan-Gen Wang \and Chunsheng Yang}

%

% TODO FINAL: Replace with an abbreviated list of authors.
\authorrunning{F. Xing et al.}
% First names are abbreviated in the running head.
% If there are more than two authors, 'et al.' is used.

% TODO FINAL: Replace with your institution list.
\institute{Guangzhou University, Guangzhou 510006, China\\
\{xfchuang, shixiaowen\}@e.gzhu.edu.cn, \{wangyg, chunsheng.yang\}@gzhu.edu.cn}
%,chunsheng.yang@nrc.gc.ca
%\url{http://www.springer.com/gp/computer-science/lncs} \and
%ABC Institute, Rupert-Karls-University Heidelberg, Heidelberg, Germany\\
%\email{\{abc,lncs\}@uni-heidelberg.de}

\maketitle

\begin{abstract}
Unveiling the real appearance of retouched faces to prevent malicious users from deceptive advertising and economic fraud has been an increasing concern in the era of digital economics. This article makes the first attempt to investigate the face retouching reversal (FRR) problem. We first collect an FRR dataset, named deepFRR, which contains 50,000 StyleGAN-generated high-resolution (1024$\times$1024) facial images and their corresponding retouched ones by a commercial online API. To our best knowledge, deepFRR is the first FRR dataset tailored for training the deep FRR models. Then, we propose a novel diffusion-based FRR approach (FRRffusion) for the FRR task. Our FRRffusion consists of a coarse-to-fine two-stage network: A diffusion-based Facial Morpho-Architectonic Restorer (FMAR) is constructed to generate the basic contours of low-resolution faces in the first stage, while a Transformer-based Hyperrealistic Facial Detail Generator (HFDG) is designed to create high-resolution facial details in the second stage. Tested on deepFRR, our FRRffusion surpasses the GP-UNIT and Stable Diffusion methods by a large margin in four widespread quantitative metrics. Especially, the de-retouched images by our FRRffusion are visually much closer to the raw face images than both the retouched face images and those restored by the GP-UNIT and Stable Diffusion methods in terms of qualitative evaluation with 85 subjects. These results sufficiently validate the efficacy of our work, bridging the recently-standing gap between the FRR and generic image restoration tasks. The dataset and code are available at https://github.com/GZHU-DVL/FRRffusion.

\end{abstract}

\section{Introduction}
\label{sec:intro}

Facial retouching filters and applications have seamlessly integrated into our daily lives, yet their pervasive usage exposes potential risks. A salient concern lies in malicious users' exploitation of retouching filters for the purpose of catfishing, deceptive advertising, and even economic defraud. Consequently, the reversal of retouching filters is paramount in bolstering online security and aiding in suspect identification. Additionally, the advertising industry has witnessed an escalating consumer demand for authenticity and transparency. Face retouching reversal (FRR) emerges as a new task, which might combat deceptive advertising practices while allowing consumers to access more genuine and reliable face-related product information.

Despite the pressing need for FRR, there is a gap between the FRR and existing image restoration tasks. The most relevant research direction is facial retouching detection \cite{bharati2016detecting,rathgeb2020prnu,wang2019detecting,kee2011perceptual}, which involves a binary classification problem and detects whether a face image has undergone the beautification modification or not. Very recently, Ying et al. \cite{ying2023retouchingffhq} created a large-scale dataset (RetouchingFFHQ) and developed a new approach for fine-grained face retouching detection on various retouching types and levels. However, these approaches cannot reverse the operation of retouching filters, failing to reveal the authenticity of brand advertising and legal evidence. Diffusion models \cite{ho2020denoising,sohl2015deep,song2019generative,song2020score} have shown great potential across various vision tasks, including image super-resolution \cite{li2022srdiff,saharia2022image,chung2022mr}, restoration \cite{lugmayr2022repaint,rombach2022high,saharia2022palette,kawar2022denoising,yang2023gp}, translation \cite{ozbey2023unsupervised,chung2022score,song2021solving}, editing \cite{preechakul2022diffusion,batzolis2021conditional,meng2021sdedit}, semantic segmentation \cite{xu2023open,baranchuk2021label,brempong2022denoising,graikos2022diffusion}, video generation \cite{harvey2022flexible,mei2023vidm,yang2023diffusion,zhang2022motiondiffuse}, anomaly detection \cite{wyatt2022anoddpm,bandara2022remote,han2022adbench}, and more. These successes not only reflect the versatility of diffusion models across various applications but also highlight their advantages over traditional methods in handling complex vision tasks.

In the context of FRR, diffusion models may produce a remarkable effect by referring to their successful practices in the aforementioned areas. This work aims to explore the feasibility of using diffusion models to remove face-retouching effects from retouched images. Specifically, we first utilize the Face++ API \footnote{https://api-cn.faceplusplus.com/facepp/v2/beautify.} to construct a deepFRR dataset. It comprises pre- and post-beauty enhancement face images, encompassing a broad range of characteristics such as genders, ages, and backgrounds. Then, we develop a new network architecture named FRRffusion, which aims to acquire the ability to reverse various retouching operations and restore images to their original states. Both the quantitative and qualitative evaluations demonstrate the effectiveness of our method. The major contributions of this work can be summarized as follows.
\begin{itemize}
\item[$\bullet$] We propose a new computer vision task called face retouching reversal (FRR). To handle this task, we collect an FRR dataset named deepFRR, which includes 50,000 StyleGAN-generated high-resolution face images and their corresponding retouched ones undergoing six distinct types of retouching operations.

\item[$\bullet$] We design a novel diffusion-based FRR model (FRRffusion), which consists of a diffusion-based Facial Morpho-Architectonic Restorer (FMAR) module and a Transformers-based Hyperrealistic Facial Detail Generator (HFDG) module. The former aims to generate the holistic structure of low-resolution facial images, while the latter serves to create the intricate facial components.

\item[$\bullet$] We conduct a series of comparative experiments and ablation studies to validate the effectiveness of our FRRffusion. Experimental results show that our method achieves consistently excellent performance in terms of both quantitative and qualitative evaluations, filling the gap between them encountered by existing advanced restoration approaches like GP-UNIT \cite{yang2023gp} and Stable Diffusion \cite{rombach2022high}.

\end{itemize}

\begin{figure*}[t]
  \centering
  \centerline{\includegraphics[width=12cm]{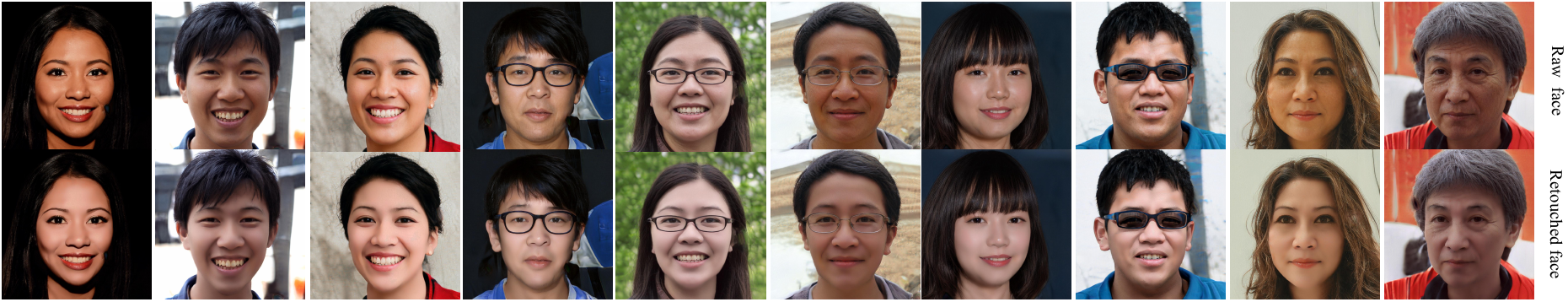}}
 \vspace{-0.5em}
\caption{Example illustration of our deepFRR. We randomly select 10 face image pairs from the deepFRR dataset. Each pair consists of a raw AI-generated face image (first row) and its corresponding retouched one (second row).}\label{fig1}
 \vspace{-1em}
\end{figure*}

\section{Related Works}
\subsection{Deepfake Detection}

Deepfake detection has emerged as a significant direction to address the misuse and proliferation of fabricated facial content. Early fake-face detection methods utilized conventional signal processing techniques and handcrafted features like facial landmarks, inconsistent lighting, and unnatural facial movements \cite{amerini2011sift,de2013exposing,fridrich2012rich,wang2014exploring,ferrara2012image,cozzolino2019noiseprint,zhou2018learning}. With the advancement of the deep neural networks (DNNs),  Generative Adversarial Networks (GANs)-based Deepfake has been developed, posing great challenges for fake-face detection. Akhtar and Dasgupta \cite{akhtar2019comparative} explored the feasibility of utilizing local feature descriptors to identify manipulated faces. Bekci et al. \cite{bekci2020cross} developed a Deepfake detection system that leverages metric learning and advanced steganalysis models to improve performance on unknown data and falsified content. Li et al. \cite{li2018ictu} investigated discrepancies in blinking patterns between Deepfake videos and genuine human subjects. Nguyen et al. \cite{nguyen2020eyebrow} employed the eyebrow region as a feature set to identify Deepfake videos. Built on random forests, Patel et al. \cite{patel2020trans} introduced Trans-DF for end-to-end Deepfake detection. Ciftci et al. \cite{ciftci2020hearts} devised a pioneering method that traces the origins of Deepfake content by analyzing biometric clues in residuals. Yang et al. \cite{yang2021msta} introduced MSTANet, which focuses on the texture features of images to identify anomalies introduced by Deepfake alterations. Recent research emphasized the increasing importance of multi-modal and multi-scale transformation in the field of Deepfake detection \cite{zhao2021multi}. Notably, Wang et al. \cite{wang2022m2tr} implemented such a transformation for Deepfake detection. Furthermore, Shao et al. \cite{shao2022detecting} addressed the detection of sequential facial manipulations in Deepfake video.

\subsection{Face Retouching Detection}

Subsequent to Deepfake detection, face retouching detection has arisen as a crucial area of facial image processing. Current methods focus on binary classification to determine whether a face image has been retouched or not. For example, a two-stage approach \cite{bharati2016detecting} was developed to improve accuracy by combining DNNs and SVM. In \cite{wang2019detecting}, variations generated by learned scripts can detect Photoshop-retouched images. Rathgeb et al. \cite{rathgeb2020prnu} utilized photo response non-uniformity for improved interpretability. In \cite{rathgeb2020differential}, multiple biometric features are used to detect retouching. Very recently, Ying et al. \cite{ying2023retouchingffhq} built a large-scale dataset (RetouchingFFHQ) and proposed a method for fine-grained face retouching detection on a wide range of retouching types and levels. However, the aforementioned methods only served to recognize the retouching filters, failing to reverse retouching operations and adapt to rapid advancements in cosmetic technology. This poses great challenges in dealing with deceptive online romance, brand authenticity, and legal evidence reliability.

\begin{figure*}[t]
  \centering
  \centerline{\includegraphics[width=12cm]{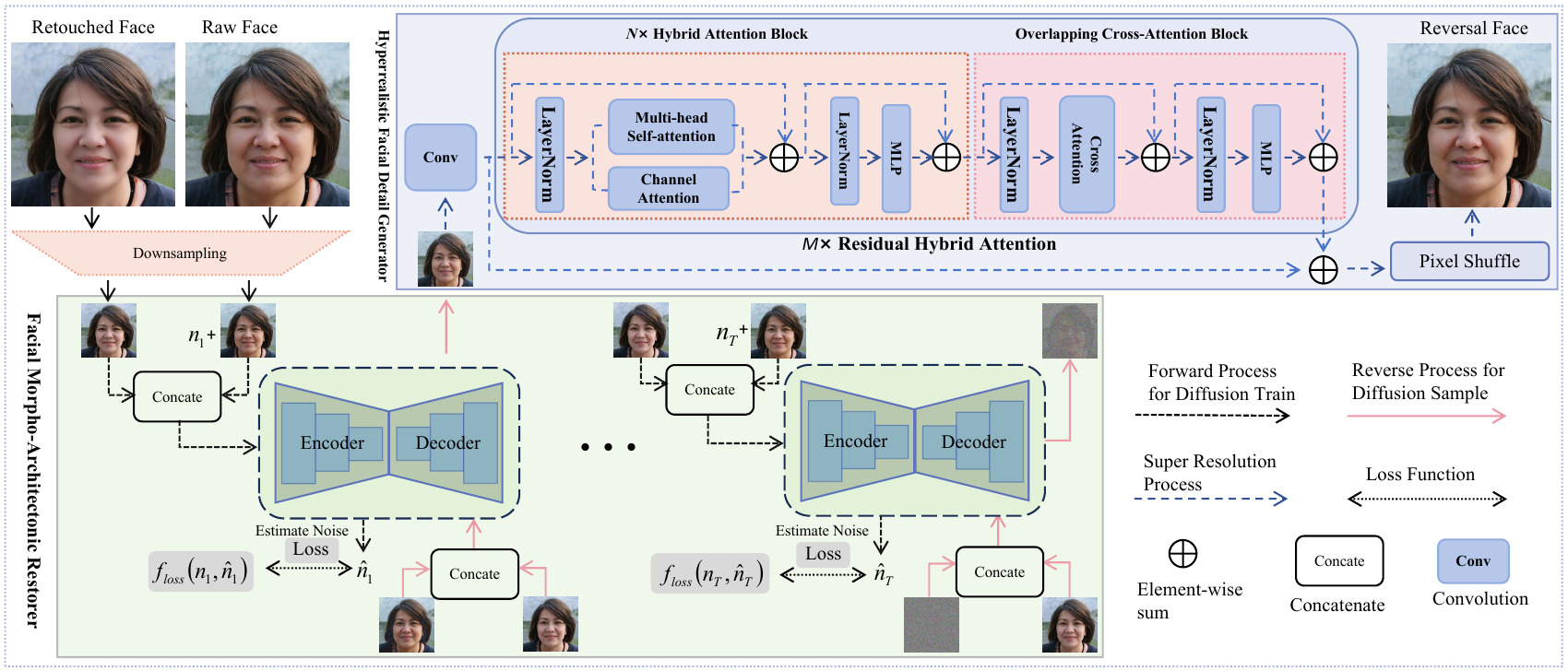}}
 \vspace{-0.5em}
\caption{Overview of the proposed FRRffusion framework. It includes Facial Morpho-Architectonic Restorer and Hyperrealistic Facial Detail Generator.} \label{framework}
 \vspace{-1em}
\end{figure*}

\section{Face Retouching Reversal Dataset}

In this section, we present the detailed process of constructing the face retouching reversal dataset (deepFRR). To avoid personal privacy and legal dispute issues, our study collects only machine-generated face images obtained from the website \footnote{http://seeprettyface.com/mydataset.html\#yellow.}. The deepFRR dataset comprises 50,000 high-resolution ($1024\times1024$) facial images that were generated using the StyleGAN model \cite{karras2020analyzing}. These face images accurately capture intricate facial features and exhibit a remarkable level of realism, offering a comprehensive representation of diverse ages, genders, and backgrounds.
To construct deepFRR, we leverage a commercial online API provided by Face++ \footnote{https://api-cn.faceplusplus.com/facepp/v2/beautify.}. This API offers a wide range of operation types for retouching face appearance, including Eye Enlarging, Face Slimming, Skin Whitening, Skin Smoothing, Eyebrow Shaping, and Face Shrinking. The detailed description can refer to the technical document \footnote{https://console.faceplusplus.com.cn/documents/134252584.}. The construction process involves uploading the raw face images and configuring specific retouching types, along with retouching levels adjusted within the range of 0 to 100, where 0 and 100 correspond to the least and maximum modification levels, respectively. Considering the computational cost, we perform only one maximum level modification for all six types of face retouching operations. As a result, we obtain the deepFRR dataset consisting of 50,000 pairs of face images, as shown in Fig. \ref{fig1}. This dataset provides valuable resources for researchers and practitioners to develop and evaluate face-retouching-related algorithms and models.

It is worth emphasizing that our deepFRR dataset significantly differs from RetouchingFFHQR \cite{ying2023retouchingffhq} by these facts. Firstly, our dataset includes six types of retouching operations with excellent retouching effects under Face++ Version 2.0. And, all of the 50,000 samples have undergone six mixed operations with the strongest level (100). By contrast, RetouchingFFHQR contains four types of retouching operations with a modest retouching effect where the best one is from Face++ Version 1.0 \footnote{https://www.faceplusplus.com.cn/sdk/facebeautify/.}. Besides, an overwhelming majority of RetouchingFFHQR involves only a single retouching operation with slight levels (i.e. 30 and 60), which may fit the retouching detection well, however, is not suitable for deceptive use of retouching face products where the retouching effect must be tuned to the maximum. This shows that our dataset is more challenging and applicable to the FRR task. Secondly, our dataset is AI-generated without visual differences from real faces, avoiding unnecessary personal privacy and portrait violations. As a result, our dataset can be easily extended in size and be widely publicized to users. However, RetouchingFFHQR consists of the real faces of users' pictures captured in the real world, acting as a supplement to our dataset when testing in real-world scenarios but being limited to widespread usage due to privacy protection problems.

\section{Proposed FRRffusion}
\label{sec:blind}
In this section, we introduce the technical details of the proposed method.  We first briefly overview the whole FRRffusion framework and then describe its two core network modules in detail, including Facial Morpho-Architectonic Restorer (FMAR) and Hyperrealistic Facial Detail Generator (HFDG). We utilize bold lowercase letters to denote (column) vectors and bold or special uppercase letters to denote matrices.

\subsection{Framework Overview}
To deal with the computational challenge of using high-resolution images to train the diffusion model, we propose to downsample the input images to obtain low-resolution samples for training. Then, we exploit the super-resolution technique to alleviate the quality degradation in FRR related to low-resolution image diffusion. The proposed FRRffusion method is a Transformer- and diffusion-based framework, as shown in Fig. \ref{framework}, which consists of a two-stage network. In the first stage, downsampling is applied to the raw face images, and our diffusion-based FMAR module generates the basic contour of the face image, capturing the global facial structure and morphology. In the second stage, our Transformer-based HFDG module further processes the facial images from the first stage, resulting in highly realistic super-resolution face images.
% HFDG focuses on enhancing the details and realism of the FRR images.

\subsection{Facial Morpho-Architectonic Restorer}
In this subsection, we provide a detailed description of our designed FMAR. The FMAR is built upon the Denoising Diffusion Probabilistic Model (DDPM) \cite{ho2020denoising}. To make this paper self-contained, we provide a brief introduction to the mathematical foundation of the DDPM.

The training procedure of DDPM comprises two essential steps: forward noising and backward denoising. In the forward process, noise is systematically added to the original sample $x_0$ with a gradually increasing diffusion rate $\beta_t$ ($\beta_t\in [0.0001, 0.02]$, $t\in [1, T]$, where $T$ is the total number of diffusion steps). This progressive noise addition produces a sequence of noisy images $x_{1:T}$. The resultant noisy images gradually converge towards a standard Gaussian distribution.

Mathematically, we can represent the forward diffusion process as a Markov chain by
\begin{align}
    \label{eq_q(x_{1:T}|x_0)}
    q(x_{1:T} | x_0) := \prod \limits_{t=1}^T q(x_t|x_{t-1}),\quad \quad \quad \quad \\
    q(x_t | x_{t-1}) := \mathcal{N}(x_t; \sqrt{(1-\beta_t)}x_{t-1}, \beta_t \textbf{I}). \quad
\end{align}
Here, we denote the current time step of the diffusion process as $t\in [1, T]$. To simplify the notation, we introduce $\alpha_t := 1 - \beta_t$ and $\bar{\alpha}_t := \prod_{i=1}^t \alpha_i$. With these definitions, we can express the probability distribution $q(x_t|x_0)$ as follows:
\begin{equation}
    \label{eq_q_x_t|x_0}
    q(x_t|x_0) = \mathcal{N}(x_t;\sqrt{\bar{\alpha}_t}x_0, (1-\bar{\alpha}_t)\textbf{I}).
\end{equation}
For a given raw image $x_0$, the sampled value $x_t$ at time step $t$ can be written by
\begin{equation}
    \label{eq_x_t}
    x_t = \sqrt{\bar{\alpha}_t} x_0 + \sqrt{1-\bar{\alpha}_t} \epsilon,\quad \epsilon \sim \mathcal{N}(0, \textbf{I}).
\end{equation}
Based on the computation of $x_t$, it can be observed that as the total number of diffusion steps $T$ increases, the value of $\sqrt{\bar{\alpha}_t}$ approaches zero. Consequently, $x_t$ will converge towards $\epsilon$.

In the denoising process, solving the posterior distribution $p(x_{t-1}|x_t)$ poses a challenge. To address this, we employ a neural network, denoted as $\theta$, to approximate this distribution. The predicted distribution is denoted as $p_\theta(x_{t-1}|x_t)$. Assuming that the mean and variance of $p_\theta(x_{t-1}|x_t)$ are $\mu_\theta(x_t, t)$ and $\sigma_\theta(x_t, t)$, respectively, we can express $p_\theta(x_{t-1}|x_t)$ as follows:
\begin{equation}
    \label{eq_p_theta(x_t-1|x_t)}
    p_\theta(x_{t-1}|x_t) = \mathcal{N}(x_{t-1}; \mu_\theta(x_t, t), \sigma_\theta(x_t, t)).  
\end{equation}
In the forward diffusion process, we can deduce that given $x_0$ and $x_t$, the distribution of $p(x_{t-1}|x_t, x_0)$ can be expressed as follows:
\begin{equation}
    \label{eq_p(x_t-1|x_t,x_0)}
    p(x_{t-1}|x_t, x_0) = \mathcal{N}(x_{t-1};\mu_t(x_t, x_0), \sigma_t),
\end{equation}
where $\mu_t(x_t, x_0) = \frac{\sqrt{\alpha}_t(1-\bar{\alpha}_{t-1})}{1-\bar{\alpha}_t} x_t + \frac{\sqrt{\bar{\alpha}_{t-1}}(1-\alpha_t)}{1-\bar{\alpha}_t} x_0$ and $\sigma_t = \frac{(1-\bar{\alpha}_{t-1})(1-\alpha_t)}{1-\bar{\alpha}_t}$.
By working out $x_0$ from Equation (\ref{eq_x_t}) and subsequently substituting it into $\mu_t(x_t, x_0)$, we can obtain the following expression:
\begin{equation}
    \label{eq_mu_t(x_t, t)}
    \mu_t(x_t, t) = \frac{1}{\sqrt{\alpha}_t}(x_t - \frac{1-\alpha_t}{\sqrt{1-\bar{\alpha}_t}} \epsilon).
\end{equation}
Given that the variance $\sigma_t$ of $p(x_{t-1}|x_t, x_0)$ is constant, predicting $p(x_{t-1}|x_t, x_0)$ is equivalent to estimating $\mu_t(x_t, t)$. Therefore, we can utilize the network model $\theta$ to parameterize $\mu_\theta(x_t, t)$. This can be achieved by
\begin{equation}
    \label{eq_mu_theta(x_t, t)}
    \mu_\theta(x_t, t) = \frac{1}{\sqrt{\alpha}_t}(x_t - \frac{1-\alpha_t}{\sqrt{1-\bar{\alpha}_t}} \epsilon_\theta(x_t, t)),
\end{equation}
where $\epsilon_\theta(x_t, t)$ represents the predicted value of the added noise at time step $t$. During the training process, the objective is to minimize the discrepancy between the predicted distribution $p_\theta(x_{t-1}|x_t)$ and the posterior distribution $p(x_{t-1}|x_t, x_0)$. This can be achieved by minimizing the expected squared difference between $\mu_t(x_t, t)$ and $\mu_\theta(x_t,t)$ over the variables $t$, $x_0$, and $\epsilon$. 
A simple and effective loss function for this minimization can be formulated as follows:
\begin{equation}
    \label{eq_loss}
    L_{simple} = \mathbb{E}_{t,x_0,\epsilon}[||\epsilon-\epsilon_\theta(x_t,t)||^2].
\end{equation}

Once the network model ($\theta$) is well-trained, it enables us to predict the noise introduced at each time step $t$. The time step $t$ starts from $1$ and gradually increases until it reaches the maximum value $T$. As $T$ becomes sufficiently large, the variable $x_T$ at time step $T$ will follow a standard Gaussian distribution. 
During the sampling process, we sample a pure noise term (denoted as $n_T$) from the standard Gaussian distribution. This noise term is then input into the well-trained diffusion model. Given the input of $n_t$ and the time step $t$, the denoised noise term $n_{t-1}$ in one step can be expressed as follows:
\begin{equation}
    \label{eq_x_t-1}
    n_{t-1} = \frac{1}{\sqrt{\alpha}_t}(n_t - \frac{1-\alpha_t}{1-\sqrt{\bar{\alpha}_t}}(\epsilon_\theta(n_t, t))) + \sigma_t z,\,\, z\sim \mathcal{N}(0, \textbf{I}).
\end{equation}
According to Equation (\ref{eq_x_t-1}), the noise introduced during image degradation can be gradually removed from $n_T$ until a target image ($n_0$) is obtained.

Our diffusion-based FMAR differs from the DDPM in both the training and sampling processes. To begin with, we require a paired dataset $\{(x_0^i, y_0^i)\}, i=0,1,...,S$ for training, where $S$ represents the dataset size. In our work, $y_0^i$ and $x_0^i$ correspond to the $i$-th retouched face image and its corresponding raw one, respectively. For the sake of simplicity, we will use the sample ($x_0, y_0$) to denote an arbitrary training sample ($x_0^i, y_0^i$) in the following text.
During the training process, apart from inputting the noisy face image ($x_t$), we also input the retouched face image ($y_0$) at each time step $t$ as a supervisory condition. This condition serves as a guide for the diffusion model to generate the FRR images.

After incorporating the conditional input $y_0$, we can define the posterior probability of our FRR diffusion model $\theta_{FRR}$ as
\begin{equation}
\label{eq_p_theta(x_t-1|x_t, y_0)}
\begin{split}
    p_{\theta_{FRR}}(x_{t-1}|x_t, y_0)= \mathcal{N}(x_{t-1}; \mu_{\theta_{FRR}}(x_t,t,y_0), \sigma_{\theta_{FRR}}(x_t,t,y_0)). 
\end{split}
\end{equation}
Accordingly, the mean of the predicted noise can be written as 
\begin{equation}
\label{eq_mu_theta(x_t, t, y_0)}
\begin{split}
    \mu_{\theta_{FRR}}(x_t, t, y_0) =\frac{1}{\sqrt{\alpha}_t}(x_t - \frac{1-\alpha_t}{\sqrt{1-\bar{\alpha}_t}} \epsilon_{\theta_{FRR}}(x_t, t, y_0)).
\end{split}
\end{equation}
Meanwhile, our loss function $L_{simple}^{FRR}$ can be defined as
\begin{equation}
    \label{eq_loss_y0}
     L_{simple}^{FRR} = \mathbb{E}_{t,x_0,y_0,\epsilon}[||\epsilon-\epsilon_{\theta_{FRR}}(x_t,t,y_0)||^2].
\end{equation}
Algorithm 1 and Algorithm 2 illustrate our training and sampling processes, respectively.

\begin{algorithm}[t]
\caption{Training a denoising model $\theta_{FRR}$}
\label{alg:algorithm1}
% \textbf{Parameter}: Optional list of parameters\\
% \textbf{Output}: Your algorithm's output
\begin{algorithmic}[1]
\STATE \textbf{Repeat} \\
\STATE  $(x_0, y_0) \sim q(x_0^i, y_0^i)$ \\
\STATE  $t \sim$ Uniform$(\{1, \ldots, T\})$ \\
\STATE  $\epsilon  \sim \mathcal{N}(0, \textbf{I})$ \\
\STATE  Take gradient descent step on \\
\quad $\nabla_\theta|| \epsilon - \epsilon_{\theta_{FRR}}(x_0, t, y_0)||^2$\\
\STATE \textbf{Until} converged
\end{algorithmic}
\end{algorithm}

\begin{algorithm}[t]
\caption{Sampling for condition $y_0$}
\label{alg:algorithm2}
% \textbf{Parameter}: Optional list of parameters\\
% \textbf{Output}: Your algorithm's output
\begin{algorithmic}[1]
\STATE  \textbf{Sample} $x_T \sim \mathcal{N}(0, \textbf{I})$ and $y_0$ 
\FOR{$t=T, \ldots, 1$}
    \STATE $z \sim \mathcal{N}(\mathbf{0}, \mathbf{I})$ if $t > 1$, else $z = \mathbf{0}$
    \STATE $x_{t-1} = \frac{1}{\sqrt{\alpha}_t}(x_t - \frac{1-\alpha_t}{1-\sqrt{\bar{\alpha}_t}}(\epsilon_{\theta_{FRR}}(x_t, t, y_0))) + \sigma_t z$
\ENDFOR
\STATE \textbf{Return} $x_0$
\end{algorithmic}
\end{algorithm}

\subsection{Hyperrealistic Facial Detail Generator}
In this subsection, we provide a detailed description of the proposed HFDG. Our HFDG is built upon the image super-resolution model HAT-B \cite{chen2023activating}. In the second stage, we acquire a low-resolution FRR image by performing the FMAR module in the first stage. Subsequently, we employ the HFDG module to increase the resolution and augment the finer details of the low-resolution image, ultimately attaining the desired high-quality super-resolution FRR image. 

For a given low-resolution input $x_0$, we first exploit one convolution layer to extract the shallow feature $f_0$, i.e., $f_0=conv(x_0)$. The shallow feature $f_0$ captures the low-level characteristics, such as edges, textures, and simple patterns. Then, the whole process of the Hybrid Attention Block (HAB) is computed as follows:
\begin{equation}
x_N = MSA(LN(f_0)) + \alpha CA(LN(f_0)) + f_0,
\end{equation}
\begin{equation}
x_M = MLP(LN(x_N)) + x_N,
\end{equation}
where $x_N$ denotes the intermediate features obtained after applying the multi-head self-attention (MSA) and channel attention (CA) mechanisms to the input feature $f_0$. The LayerNorm (LN) layer is applied to the input $f_0$ before feeding it into the MSA and CA modules. The MSA module captures long-range dependencies and spatial relationships within the input feature maps, while the CA module focuses on channel-wise feature refinement. A small constant $\alpha$ is multiplied by the output of CA and controls the trade-off between the two attention mechanisms. The intermediate features $x_N$ are then processed by a multi-layer perceptron (MLP) module, preceded by another LN operation. Finally, the output $x_M$ of the HAB is obtained by adding the MLP's output to the intermediate features $x_N$, enabling a residual connection.

After passing through $N$ HAB modules, $x_M$ will enter the overlapping cross-attention block (OCAB)  module for further processing. The computation process can be represented as follows:
\begin{equation}
x_C = Cross\_Attention(LN(x_M)) + x_M,
\end{equation}
\begin{equation}
y = MLP(LN(x_C)) + x_C,
\end{equation}
where $Cross\_Attention(\cdot)$ function captures the long-range dependencies and contextual information within the input features, enabling the model to effectively integrate information across different spatial locations. The output of the cross-attention operation, $x_C$, is then combined with the original input $x_M$ through a residual connection. The output of the MLP is added back to $x_C$ using another residual connection, yielding the final output $y$ of the OCAB module. After passing the input features $y$ through $M$ Residual Hybrid Attention (RHA) modules, the extraction of deep features has been accomplished.

The output feature $y$ is added to the shallow features $f_0$ extracted from the initial convolutional layer. This combined representation is passed through a Pixel Shuffle module to obtain the reconstructed high-resolution face image. The overall process can be formulated as:

\begin{equation}
x_{HR} = Pixel\_Shuffle(y+f_0),
\end{equation}
where $Pixel\_Shuffle(\cdot)$ denotes the operation of reconstructing the image, $x_{HR}$ denotes the reconstructed high-resolution image. Interested readers may refer to the seminal work of HAT-B \cite{chen2023activating} for a comprehensive and detailed description.

\section{Experimental Results and Analyses}
In this section, we conducted comprehensive experiments to evaluate the effectiveness of our FRRffusion model. We first outline the experimental setup in Subsection \ref{Experimental Setup}. Then, we evaluate the performance of FRRffusion on individual datasets in Subsection \ref{Performance}, as well as cross-datasets in Subsection \ref{Performance1}. Furthermore, we adopt four top-performance face recognition models for quantitative evaluation, as shown in Subsection \ref{Performance2}.  Additionally, we present qualitative results and analyses for the proposed FRRffusion in Subsection \ref{Qualitative}. Finally, we conduct ablation studies in Subsection \ref{Ablation} to verify the contribution of each component to FRRffusion.

\begin{table*}[tp]
\centering
\caption{\small SSIM and PSNR Performances of Comparison Methods on Individual Datasets.} \label{tab1}
\begin{adjustbox}{width=\textwidth}
\begin{tabular}{l:lcc|cc|cc|cc}
\Xhline{1pt} %
\multicolumn{2}{c}{Type/ } &\multicolumn{8}{c}{Intra-dataset Test sets}        \\
\hdashline%
\multicolumn{2}{c}{Testing sets/} &\multicolumn{2}{c|}{deepFRR} &\multicolumn{2}{c|}{RetouchingFFHQ-1} &\multicolumn{2}{c|}{RetouchingFFHQ-2} &\multicolumn{2}{c}{RetouchingFFHQ-3}  \\ \hline
   Size & Methods                    &  SSIM & PSNR           &  SSIM & PSNR      &  SSIM & PSNR        &  SSIM & PSNR              \\    \hline
\multirow{4}{*}{\makecell[l]{512$\times$512}}
                        &GP-UNIT \cite{yang2023gp}      & 0.787 & 29.825         & 0.632 & 29.323          & 0.649  & 29.646     &0.649  & 29.588                          \\
                        &Stable Diffusion \cite{rombach2022high}     & 0.726 & 29.734        & 0.677 & 29.779          & 0.681     & 29.633        & 0.682    & 29.776                            \\
                        &\bf{Our FRRffusion}     & \textbf{0.884} & \textbf{33.049}      & \textbf{0.850}  & \textbf{30.035}     & \textbf{0.847} & \textbf{30.955}    &\textbf{0.887} &\textbf{33.331}    
                        \\   \cdashline{2-10}  
                        & Raw and Retouched     & 0.876 & 31.939   & 0.919 & 36.280      & 0.948 & 34.208      & 0.917 & 36.262  \\                  
\Xhline{1pt} %
\end{tabular}
\end{adjustbox}
\end{table*}

\begin{table*}[tp]
\centering
\caption{\small VGGS and CLIPS Performances of Comparison Methods on Individual Datasets.} \label{tab2}
\begin{adjustbox}{width=\textwidth}
\begin{tabular}{l:lcc|cc|cc|cc}
\Xhline{1pt} %
\multicolumn{2}{c}{Type/ } &\multicolumn{8}{c}{Intra-dataset Test sets}        \\
\hdashline%
\multicolumn{2}{c}{Testing sets/} &\multicolumn{2}{c|}{deepFRR} &\multicolumn{2}{c|}{RetouchingFFHQ-1} &\multicolumn{2}{c|}{RetouchingFFHQ-2} &\multicolumn{2}{c}{RetouchingFFHQ-3}  \\ \hline
   Size & Methods        &  VGGS & CLIPS  &  VGGS & CLIPS    &  VGGS & CLIPS    &  VGGS & CLIPS               \\    \hline
\multirow{4}{*}{\makecell[l]{512$\times$512}}
                        &GP-UNIT \cite{yang2023gp}      & 0.917 & 0.961         & 0.898 & 0.935          & 0.912  & 0.945     &0.915  & 0.943                          \\
                        &Stable Diffusion \cite{rombach2022high}     & 0.810 & 0.937        & 0.816 & 0.958    & 0.814 & 0.958         & 0.802 & 0.958                            \\
                        &\bf{Our FRRffusion}     & \textbf{0.991} & \textbf{0.973}     & \textbf{0.947} & \textbf{0.974}   & \textbf{0.920}  & \textbf{0.962}     & \textbf{0.994} & \textbf{0.979}        
                        \\   \cdashline{2-10}  
                        & Raw and Retouched     & 0.936 & 0.969  & 0.989 & 0.996      & 0.988 & 0.993      & 0.988 & 0.995  \\                  
\Xhline{1pt} %
\end{tabular}
\end{adjustbox}
\end{table*}

\subsection{Experimental Setup}\label{Experimental Setup}
We introduce the experimental settings in this subsection, including dataset description and baseline methods, implementation and parameter details, and performance metrics.

\subsubsection{Dataset Description and Baseline Methods.} 
In this study, we employ four datasets: our created deepFRR dataset, the RetouchingFFHQR-1, -2, and -3 datasets which are subsets extracted from the comprehensive RetouchingFFHQR dataset \cite{ying2023retouchingffhq}. The deepFRR serves as a fundamental benchmark dataset for the FRR task. The RetouchingFFHQR-1, RetouchingFFHQR-2, and RetouchingFF- HQR-3 datasets are responsible for generalization testing. These three subsets respectively correspond to the Alibaba \footnote{https://help.aliyun.com/document\_detail/159210.html.},  Tencent \footnote{https://cloud.tencent.com/document/product/1172/40715.}, and Megvii \footnote{https://www.faceplusplus.com.cn/sdk/facebeautify/.} API operations, where only one operation of four types of retouching (i.e. Skin Smoothing, Face Whitening, Face Lifting, and Eye Enlarging) is applied. Specifically, RetouchingFF- HQR-1, RetouchingFFHQR-2, and RetouchingFFHQR-3 include 36,000, 192,000, and 57,304 retouched face images with one of three levels (slight: 30, medium: 60, and heavy: 90). Note that the FRR problem is first investigated in this work. To our best effort, we choose two top-performance image generation methods for comparative evaluation, which are GP-UNIT \cite{yang2023gp} and Stable Diffusion \cite{rombach2022high}.

\subsubsection{Implementation and Parameter Details.} Our network is built on the PyTorch framework \cite{paszke2019pytorch} and trained using six Tesla 4090 GPU cards. In the first stage, we set $T$ to 1000. The downsampled images inputted into the diffusion model are of size $128\times128$, and the network is trained for 1,500,000 iterations which shows a good enough convergence. In the second stage, we fine-tune HAT-B \cite{chen2023activating} where all parameters are kept the same as given in the published paper. The deepFRR dataset is split into training and testing sets in an 8:2 ratio. Finally, our FRRffusion model occupies 125.003 MB parameters with 88.133 GFLOPS computation costs.

\subsubsection{Performance Metrics.} To evaluate the performance of the proposed FRRffusion, we employ four common metrics: Peak Signal-to-Noise Ratio (PSNR), Structural Similarity Index (SSIM) \cite{wang2004image}, VGG score (VGGS) \cite{simonyan2015a}, and CLIP score (CLIPS)\cite{radford2021learning}. SSIM measures image structural similarity by considering the brightness, contrast, and structural information. PSNR quantifies image reconstruction quality by measuring the Mean Squared Error (MSE) between the original and reconstructed images, which is particularly suitable for objective pixel-level distortion evaluation. VGGS and CLIPS evaluate the perceptual similarity between the original face image and the de-retouched one. VGGS is computed using the VGG network trained on the VGGface2 dataset, providing effective similarity quantification focused on the local face region. CLIPS is computed using the CLIP network trained on a large-scale image-text paired dataset, enabling superior similarity assessment for the overall global image.

\begin{table*}[tp]
\centering
\caption{Performance Evaluation on Cross-Datasets.}\label{tab3}
%\begin{adjustbox}{width=\textwidth}
\begin{tabular}{l:cc cc cc}
\Xhline{1pt} %
\multicolumn{2}{c}{Training sets/ } &\multicolumn{4}{c}{deepFRR}       \\
\hdashline%
\multicolumn{1}{c}{Testing sets/ }  & \multicolumn{2}{c}{RetouchingFFHQ-1}   & \multicolumn{2}{c}{RetouchingFFHQ-2}       & \multicolumn{2}{c}{RetouchingFFHQ-3}  \\  \hline
   Input Size &                     SSIM & PSNR  &  SSIM & PSNR      &  SSIM & PSNR                     \\           \hline
\multirow{3}{*}{512$\times$512}
                                        & 0.843&26.156   & 0.839&23.534       & 0.879  & 30.414     \\ \cdashline{2-7}
                                        &  VGGS & CLIPS  &  VGGS & CLIPS     &  VGGS & CLIPS    \\ \cline{2-7}
                                        & 0.977&0.972    & 0.975&0.969       & 0.989  & 0.978\\   \hline\hline
                                        
\multicolumn{2}{c}{Training sets/ } &\multicolumn{4}{c}{RetouchingFFHQ-1}       \\
\hdashline%
\multicolumn{1}{c}{Testing sets/ }  & \multicolumn{2}{c}{deepFRR}   & \multicolumn{2}{c}{RetouchingFFHQ-2}       & \multicolumn{2}{c}{RetouchingFFHQ-3}  \\  \hline
   Input Size &                     SSIM & PSNR  &  SSIM & PSNR      &  SSIM & PSNR                     \\           \hline
\multirow{3}{*}{512$\times$512}
                                        & 0.826&23.745  & 0.839&23.747   & 0.845  & 23.974     \\ \cline{2-7}
                                        &  VGGS & CLIPS  &  VGGS & CLIPS         &  VGGS & CLIPS \\ \cline{2-7}  
                                        & 0.976&0.966    & 0.969&0.972       & 0.976& 0.972  \\   \hline\hline
                                        
\multicolumn{2}{c}{Training sets/ } &\multicolumn{4}{c}{RetouchingFFHQ-2}       \\
\hdashline%
\multicolumn{1}{c}{Testing sets/ }  & \multicolumn{2}{c}{deepFRR}   & \multicolumn{2}{c}{RetouchingFFHQ-1}       & \multicolumn{2}{c}{RetouchingFFHQ-3}  \\  \hline
   Input Size &                     SSIM & PSNR  &  SSIM & PSNR      &  SSIM & PSNR                     \\           \hline
\multirow{3}{*}{512$\times$512}
                                        & 0.825&23.476  & 0.831&24.140   & 0.829  & 23.421     \\ \cline{2-7}
                                        &  VGGS & CLIPS  &  VGGS & CLIPS         &  VGGS & CLIPS  \\ \cline{2-7}  
                                        & 0.975&0.970  & 0.969&0.965   & 0.969  & 0.968  \\  \hline\hline                    
\multicolumn{2}{c}{Training sets/ } &\multicolumn{4}{c}{RetouchingFFHQ-3}       \\
\hdashline%
\multicolumn{1}{c}{Testing sets/ }  & \multicolumn{2}{c}{deepFRR}   & \multicolumn{2}{c}{RetouchingFFHQ-1}       & \multicolumn{2}{c}{RetouchingFFHQ-2}  \\  \hline
   Input Size &                     SSIM & PSNR  &  SSIM & PSNR      &  SSIM & PSNR                     \\           \hline
\multirow{3}{*}{512$\times$512}
                                        & 0.852&28.460  & 0.877&30.355   & 0.854  & 26.225     \\ \cline{2-7}
                                        &  VGGS & CLIPS  &  VGGS & CLIPS         &  VGGS & CLIPS  \\ \cline{2-7} 
                                        & 0.977& 0.983  & 0.991&0.977   & 0.975  & 0.977  \\
\Xhline{1pt} %
\end{tabular}
%\end{adjustbox}
\end{table*}

\subsection{Quantitative Evaluation on Individual Datasets}\label{Performance}
In this subsection, we evaluate the performance of our FRRffusion and the comparison methods on the four datasets. Note that in the experiment, we resize all images of the deepFRR dataset into 512$\times$512 size for ease of computation. The experimental results are reported in Tables \ref{tab1} and \ref{tab2}, and the following interesting observations can be achieved.

\begin{itemize}
\item[$\bullet$] Firstly, our FRRffusion outperforms the other two methods by a significant margin in terms of all four metrics on all four datasets. Especially, for the deepFRR dataset, our approach respectively gains a remarkable 22\%, 11\%, 20\%, and 6\% improvement in SSIM, PSNR, VGGS, and CLIPS compared to Stable Diffusion. These impressive improvements can be attributed to the fact that our FRRffusion is specifically designed for the FRR task, while Stable Diffusion is a general generative model that lacks the targeted specificity required for FRR. Compared to GP-UNIT, our method consistently performs better in terms of the  SSIM, PSNR, VGGS, and CLIPS objective metrics. It is worth noting that GP-UNIT focuses on style transformation and lacks adequate handling of facial colorization.

\item[$\bullet$] Secondly, in comparison to the Raw and Retouched baseline, our FRRffusion still shows promising outcomes although the improvements are not higher than that of GP-UNIT and Stable Diffusion. This can be attributed to the inherent nature of FRRffusion, which is a generative model. This manner is inevitable to cause global pixel-level modification of the whole image. However, the face retouching operation focuses on local modification of facial regions while keeping the majority of the image unchanged, such as the background region. Interestingly, despite the moderate improvements in objective metrics achieved by FRRffusion over the Raw and Retouched baseline, our FRRffusion method is indeed noteworthy in terms of subjective perceptual quality. This point will be verified through the subsequent qualitative analyses in Subsection. \ref{Qualitative}.

\item[$\bullet$] Thirdly, the performance improvement of our FRRffusion on Retouching- FFHQR-3 appears much higher than those on RetouchingFFHQR-1 and RetouchingFFHQR-2. In other words, our method exhibits noticeable performance degradation on RetouchingFFHQR-1 and RetouchingFFHQR-2. This is because RetouchingFFHQR-3 is constructed by Megvii Face++ Version 1.0, while our FRRffusion is trained on deepFRR which is constructed by Megvii Face++ Version 2.0. Thus, such a similar face retouching algorithm may lead to better matching performance, posing a new challenge for the generalization ability of the designed network.

\item[$\bullet$] Finally, by comparing the Raw and Retouched baseline with GP-UNIT and Stable Diffusion, a counter-intuitive result seems to appear in these objective metrics. That is, the restored retouching faces by these two powerful techniques show a lower objective quality. This can be similarly interpreted as they perform a global modification on the whole face image, resulting in a mean large change in pixel values. But it is worth emphasizing that by carefully observing Fig. \ref{fig4}, GP-UNIT achieves at least not worse gain in perceptual quality, compared to the Raw and Retouched baseline. These results show that our work fills the gap of the FRR task between the subjective and objective similarity evaluations, which is indeed not bridged by these two popular image generation methods.
\end{itemize}

\subsection{Quantitative Evaluation on Cross-Datasets}\label{Performance1}
In this part, we conducted cross-validation experiments to evaluate the generalization performance of the proposed FRRffusion. The results are presented in Table \ref{tab3} and the following interesting observations can be made. Compared to the performance on individual datasets, experimental results tested on cross-datasets demonstrate a slight decrease, but still exhibit consistently satisfactory generalization in all four objective metrics. Especially, the performance in VGGS and CLIPS is better than that in SSIM and PSNR. This is because, compared with the SSIM and PSNR concentrated on pixel-level distortion measure, the VGGS and CLIPS metrics can better reflect the perceptual similarity between different facial images of the same identity from the face recognition perspective. Besides, we can see that the performance across the deepFRR and RetouchingFFHQ-3 datasets exceeds that of the other across-validation. This might be interpreted as these two datasets underwent a similar retouching algorithm, i.e., Megvii Face Retouching API.

\begin{figure*}[tp]
  \centering
  \begin{minipage}[b]{0.48\textwidth}
    \centering
    \includegraphics[width=\textwidth]{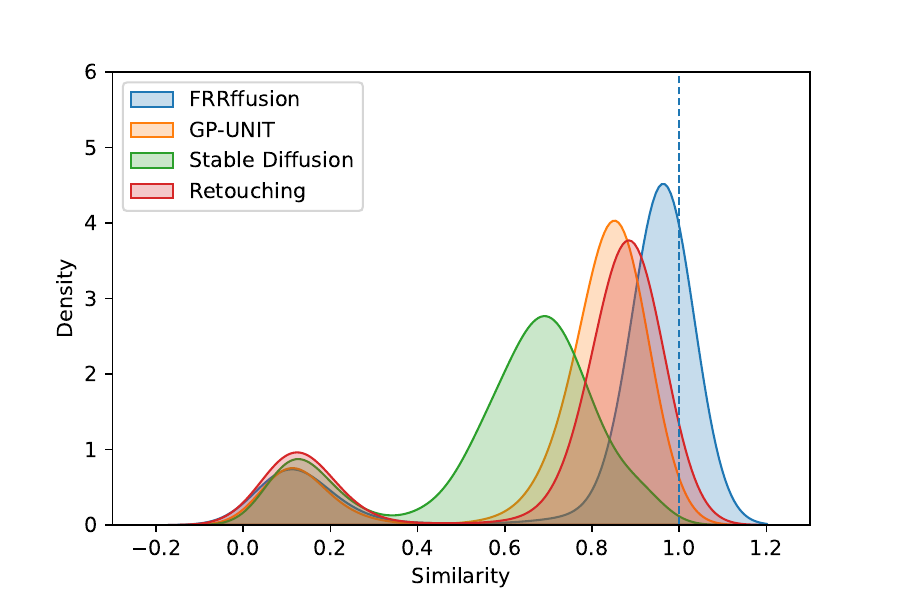}
    \subcaption*{(a) VGG-Face \cite{simonyan2015a}}
  \end{minipage}
  \hfill
  \begin{minipage}[b]{0.48\textwidth}
    \centering
    \includegraphics[width=\textwidth]{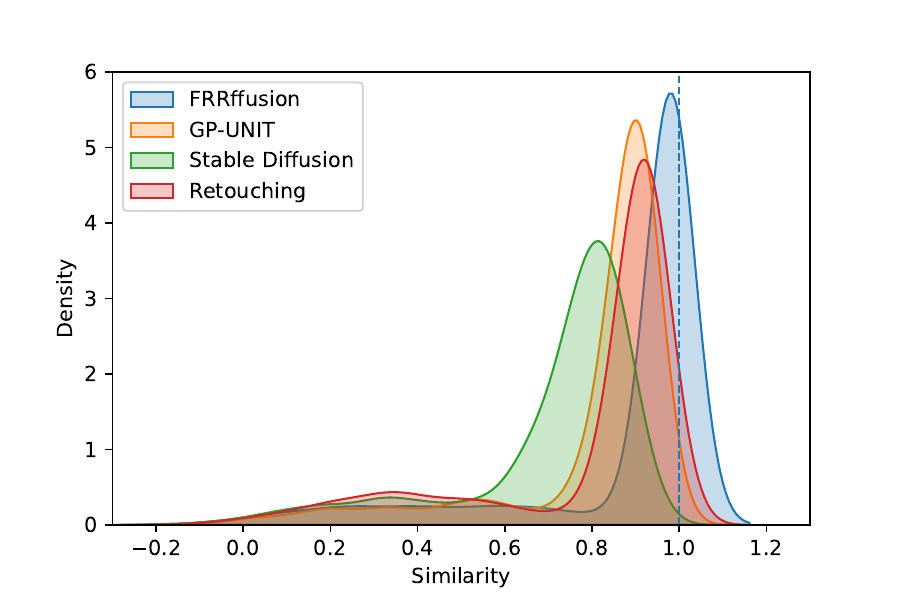}
    \subcaption*{(b) FaceNet \cite{schroff2015facenet}}
  \end{minipage}
  \hfill
  \begin{minipage}[b]{0.48\textwidth}
    \centering
    \includegraphics[width=\textwidth]{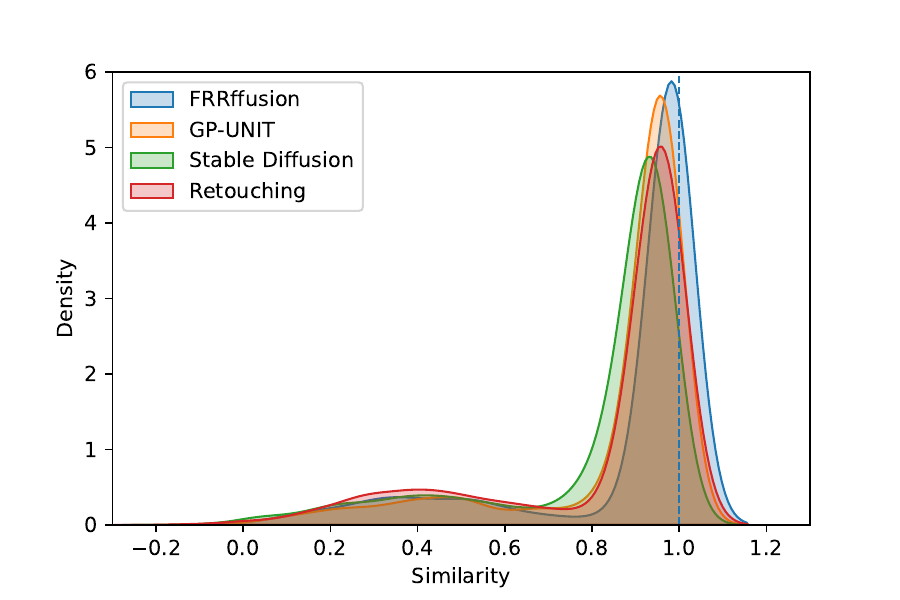}
    \subcaption*{(c) OpenFace \cite{baltruvsaitis2016openface}}
  \end{minipage}
  \hfill
  \begin{minipage}[b]{0.48\textwidth}
    \centering
    \includegraphics[width=\textwidth]{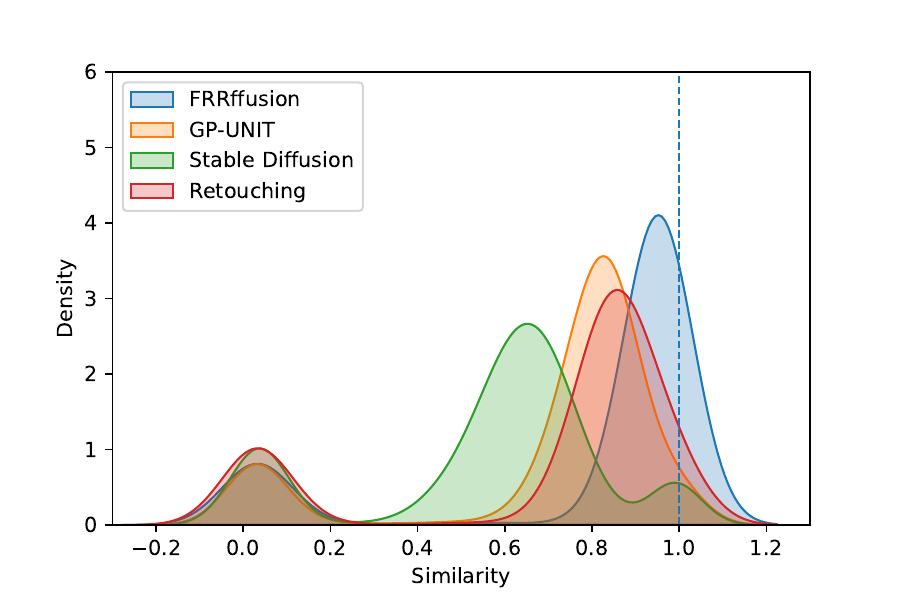}
    \subcaption*{(d) ArcFace \cite{deng2019arcface}}
  \end{minipage}
  \caption{Density distribution of the cosine similarities between the features of the raw and retouched/de-retouched face images through different face recognition models. Note that this density curve is fitted through a standard Gaussian Kernel.}
  \label{figdense}
\end{figure*}

\subsection{Quantitative Evaluation via Face Recognition Models}\label{Performance2}
To further validate the efficacy of our FRRffusion from a recognition similarity perspective, we employ four top-performance face recognition models to measure the recognition similarity between the raw and retouched/de-retouched face images, which are VGG-Face \cite{simonyan2015a}, FaceNet \cite{schroff2015facenet}, OpenFace \cite{baltruvsaitis2016openface}, and ArcFace \cite{deng2019arcface}. The results are shown in Fig. \ref{figdense}. We can see from Fig. \ref{figdense} that the Gaussian peak of our FRRffusion is closest to the reference value 1 among all the comparison methods. This shows that our FRRffusion achieves the best de-retouched effect. Besides, it is observed that the Gaussian peak of GP-UNIT is slightly farther away from the value 1 than that of the Raw and Retouched baseline (i.e. Retouching in Fig. \ref{figdense}). However, Stable Diffusion goes extremely far away. This further verifies that Stable Diffusion is incapable of carrying out the FRR tasks. We can still observe that for the FRR task, FaceNet (Fig. \ref{figdense}-b) and OpenFace (Fig. \ref{figdense}-c) obtain superior recognition performance over VGG-Face (Fig. \ref{figdense}-a) and ArcFace (Fig. \ref{figdense}-d).

\begin{figure*}[tp]
  \centering
  \centerline{\includegraphics[width=12cm]{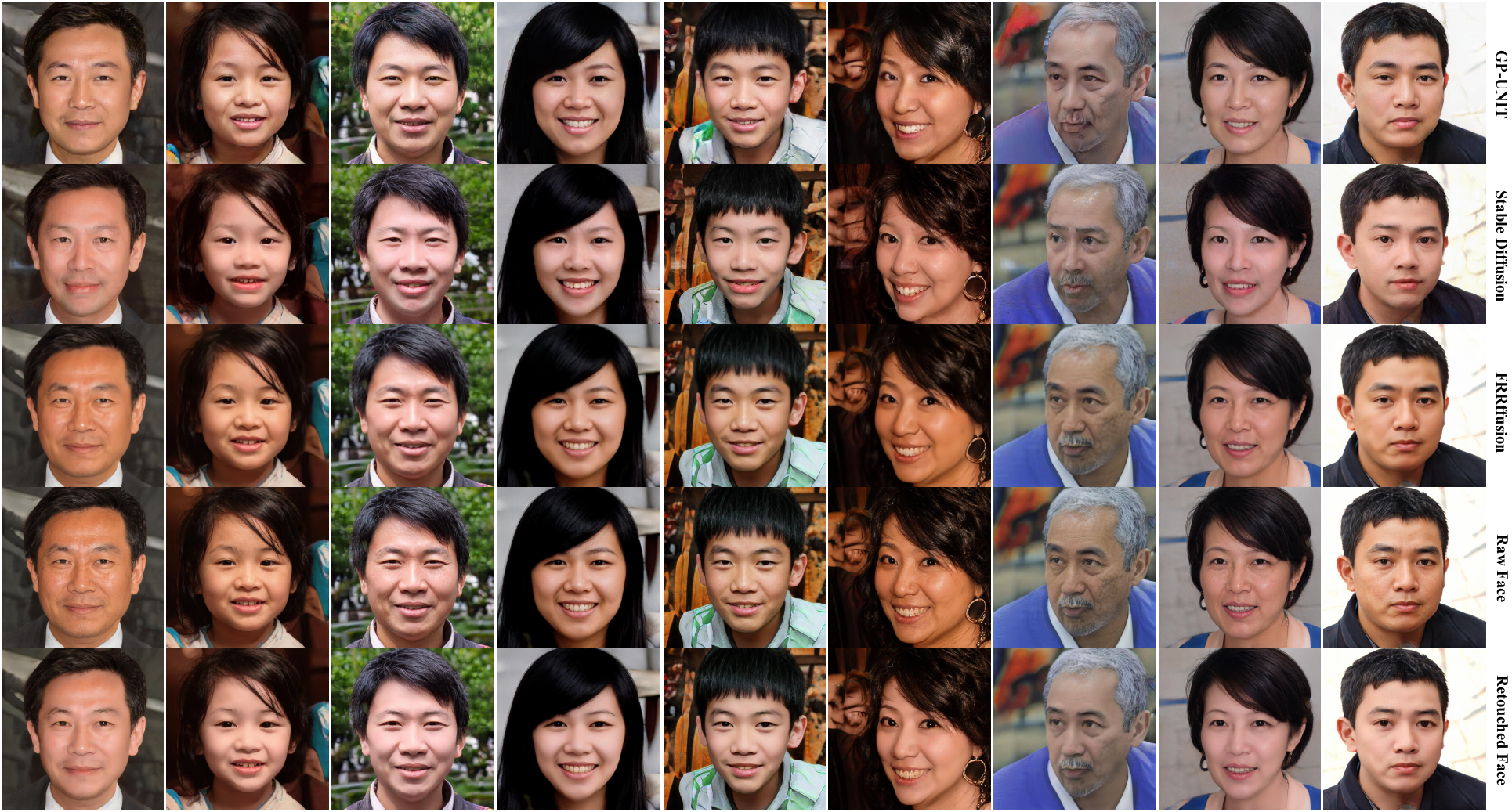}}
\caption{Qualitative comparison among different FRR methods.}\label{fig4}
\end{figure*}

\begin{figure*}[tp]
  \centering
  \includegraphics[width=10cm]{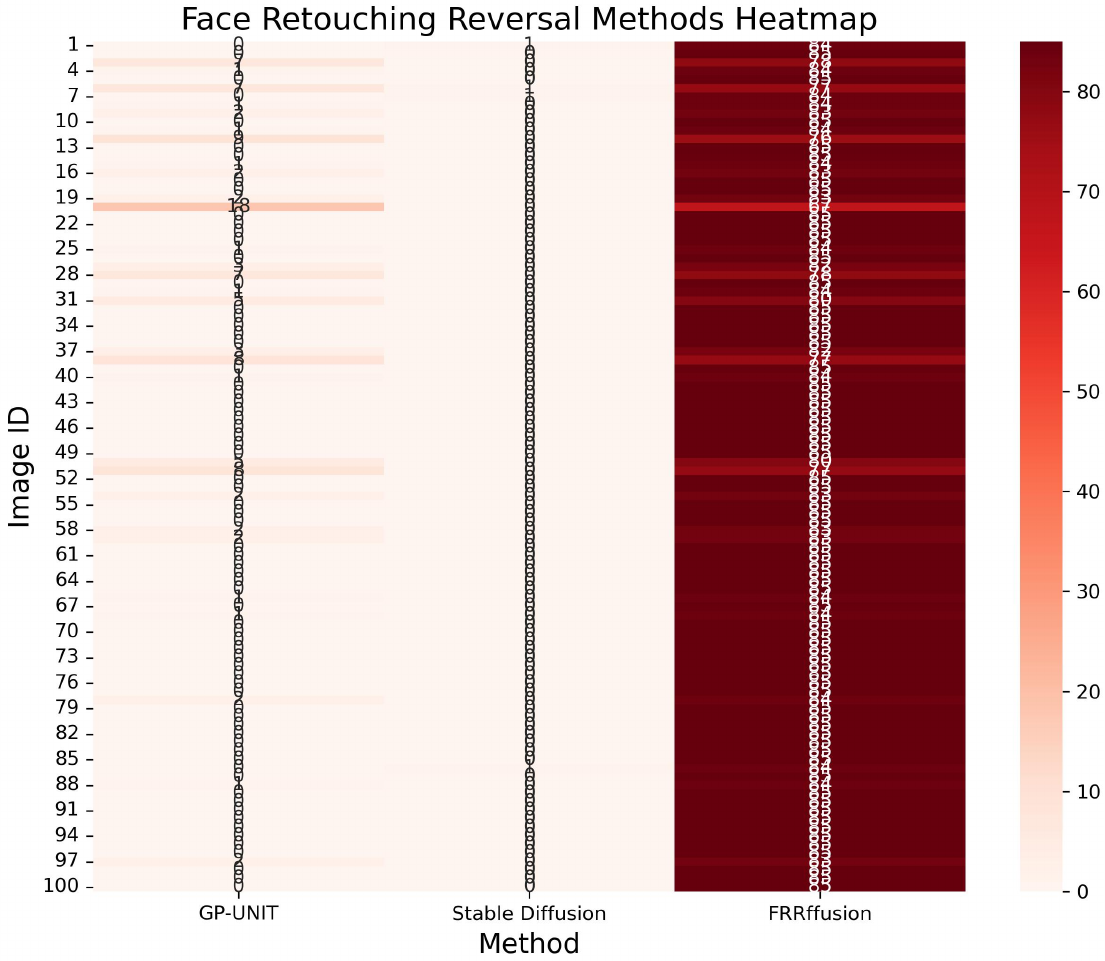}
\caption{Heatmap visualization of subjective evaluation results.}\label{fig5}
\end{figure*}

\begin{figure*}[tp]
  \centering
  \includegraphics[width=\textwidth]{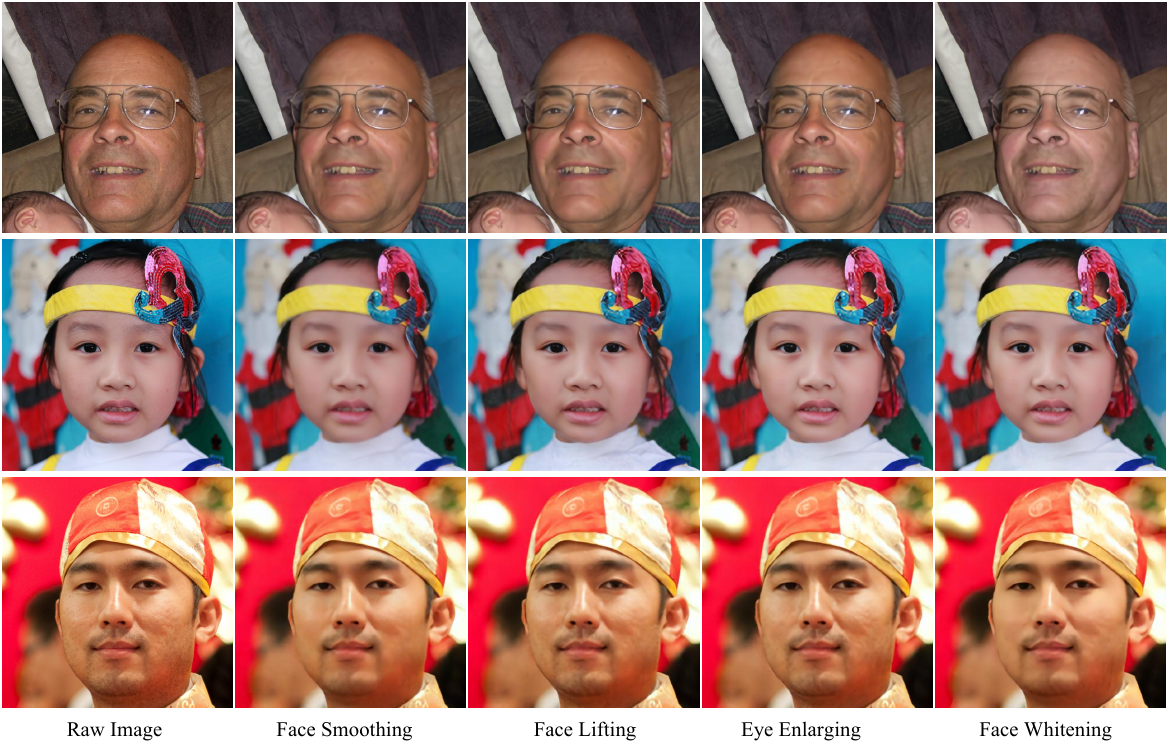}
\caption{Qualitative illustration of our FRRffusion method on FRR effects. The three images of the same column are subjected to the same single type of face retouching operations, where all images are randomly selected from RetouchingFFHQ-3.}\label{fig6}
\end{figure*}

\begin{figure*}[tp]
  \centering
  \includegraphics[width=12cm]{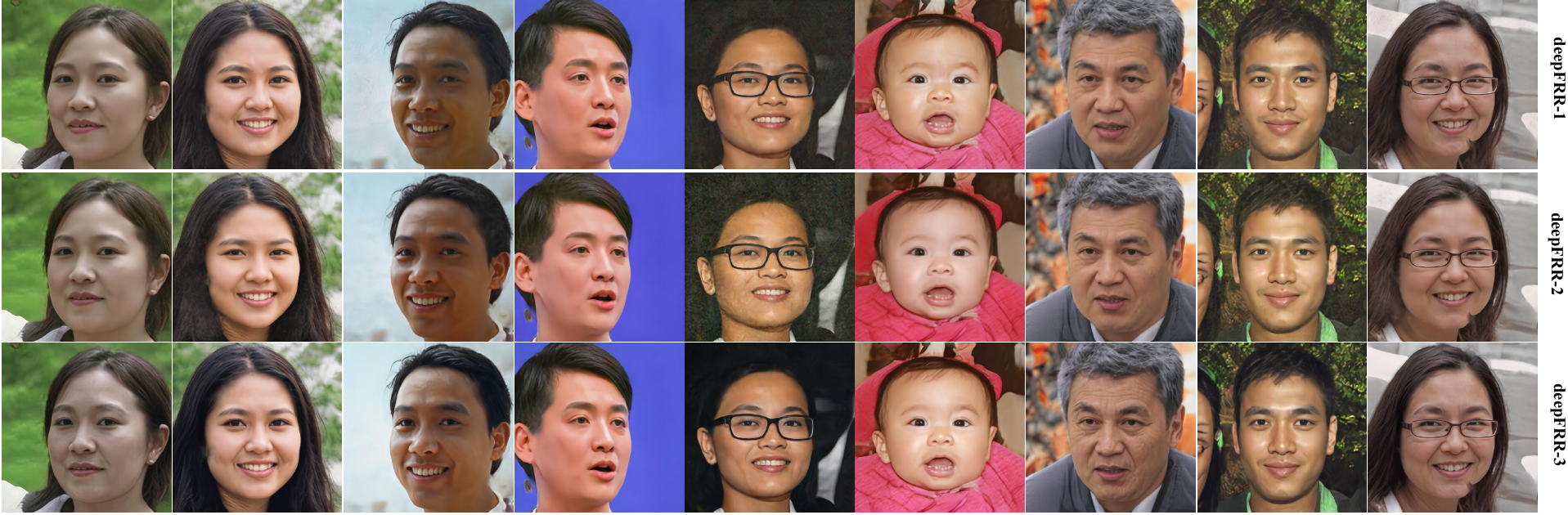}
\caption{Qualitative illustration of cross-validation on different datasets, where deepFRR-1 (first row), deepFRR-2 (second row), and deepFRR-3 (third row) show the subjective evaluation of our FRRffusion method tested on deepFRR but trained on RetouchingFFHQ-1, RetouchingFFHQ-2, and RetouchingFFHQ-3, respectively.}\label{fig7}
\end{figure*}

\subsection{Qualitative Analyses}\label{Qualitative}
In this subsection, we provide plenty of subjective results and analyses to demonstrate the effectiveness of the proposed FRRffusion. A set of samples are randomly selected from the deepFRR dataset, and their counterparts restored by the compared methods are shown in Fig.~\ref{fig4}. We can observe by zooming in on these images that the face images processed by our FRRffusion yield much higher similarity with their raw ones, compared to the retouching face images. Such higher similarity appears in various aspects including skin color, facial size, facial shape, eye size, facial details, and eyebrow shape. This indicates our method achieves a very good FRR effect.

Unfortunately, we observe that the restored face images by Stable Diffusion show larger dissimilarity with the raw faces than the retouching face images, indicating that such a powerful image generation method (Stable Diffusion) is not up to the FRR task. This finding corroborates the quantitative outcomes obtained in the earlier discussion in Subsection. \ref{Performance}. To our surprise, a counter-intuitive phenomenon occurs in GP-UNIT. That is, most faces generated by GP-UNIT are more similar to their raw ones than the retouching face images from a visual aspect. However, when we recall the quantitative evaluation shown in Subsection. \ref{Performance}, we find the four objective metrics obtained by the GP-UNIT method are worse than those of the Raw and Retouched baseline. This shows inconsistency between the subjective and objective measures, reflecting the fact that the existing methods cannot handle the FRR problem well. Promisingly, our proposed FRRffusion fills this gap, achieving consistency between the subjective and objective metrics.

Moreover, we conduct a subjective experiment on the deepFRR dataset. 100 raw facial images are randomly selected and processed using three techniques: GP-UNIT, Stable Diffusion, and our FRRffusion. Ratings are collected from 85 participants who compare the processed images and select the image with the best resemblance to the raw image. The average scores for Stable Diffusion, GP-UNIT, and FRRffusion are 0.04, 2.09, and 83.27, respectively. Note that 0.04+2.09+83.27=85.4, reflecting that 40 out of 8,500 votes are repetitively assigned to at least two images as the best resemblance. This yields a 0.47\% subjective error. Such minor errors can be neglected in subjective experiments. To better illustrate the performance comparison of the three methods, we visualize their scores on each image using a heatmap, as shown in Fig. \ref{fig5}. The heatmap provides an overview of the scores obtained by each method. It is observed that our FRRffusion achieves a remarkably high score out of 85, with only a few exceptions. The GP-UNIT method performs on few images but fails on the most, scoring 2.09. In contrast, the Stable Diffusion method performs most poorly, with a significant number of images receiving low scores, even 0.

Finally, we show the generalization ability of our method from a visual aspect. As shown in every row of Fig.~\ref{fig6}, face images with a single retouching operation restored by our FRRffusion are highly similar to the raw image. This indicates that our FRRffusion can cope with a single type of retouching operations, including Face Smoothing, Face Lifting, Eye Enlarging, and Face Whitening. What's more, in Fig.~\ref{fig7}, we showcase the cross-APIs performance where our FRRffusion is trained on different API retouching datasets and tested on the same deepFRR dataset. It is observed that the images of each column look almost the same. These two results show that our FRRffusion obtains outstanding generalization ability in both single retouching and cross-APIs retouching operations.

\subsection{Ablation Study}\label{Ablation}
In this subsection, we conduct an ablation study to investigate the contribution of each module of the proposed FRRffusion to the overall FRR performance.

\begin{table*}[tp]
\centering
\caption{\small Ablation study of different downsampling sizes. The best performances are highlighted in bold.} \label{Ablation1}
\begin{tabular}{l:l|cc|cc}
\Xhline{1pt} %
\multicolumn{2}{c}{Type/ } &\multicolumn{4}{c}{Intra-dataset Test sets}        \\
\hdashline%
\multicolumn{2}{c}{Testing sets/} &\multicolumn{4}{c}{FaceFRR}   \\ \hline
   Image Resolution & Methods                  &  SSIM & PSNR        &  VGGS & CLIPS        \\    \hline
                       32 $\times$ 32 &  HAT-B    & 0.855 & 30.535       & 0.951 & 0.976       \\
                       64 $\times$ 64 &  HAT-B    & 0.863 & 30.934       & 0.967  & 0.947      \\
                      128 $\times$ 128 &  HAT-B    & \textbf{0.884} & \textbf{33.049} &\textbf{0.973}  &\textbf{0.991}      \\             
\Xhline{1pt} %
\end{tabular}
\end{table*}

\begin{table*}[tp]
\centering
\caption{\small Ablation study of various super-resolution algorithms. The best performances are highlighted in bold.} \label{Ablation2}
\begin{tabular}{l:l|cc|cc}
\Xhline{1pt} %
\multicolumn{2}{c}{Type/ } &\multicolumn{4}{c}{Intra-dataset Test sets}        \\
\hdashline%
\multicolumn{2}{c}{Testing sets/} &\multicolumn{4}{c}{FaceFRR}   \\ \hline
   Image Resolution & Methods                          &  SSIM & PSNR        &  VGGS & CLIPS        \\    \hline
                       128$\times$ 128 &  Real-ESRGAN \cite{wang2021real}    & 0.777 & 30.075       & 0.979 & 0.971       \\
                       128$\times$ 128 &  BFRffusion \cite{chen2023towards}    & 0.764 & 30.217       & \textbf{0.982}  & 0.983      \\
                       128$\times$ 128 &  HAT-B  \cite{chen2023activating}        & \textbf{0.884} & \textbf{33.049}       &0.973  &\textbf{0.991}      \\             
\Xhline{1pt} %
\end{tabular}
\end{table*}

\subsubsection{Facial Morpho-Architectonic Restorer (FMAR).} To comprehensively assess the influence of the FMAR module on the overall performance, we resize the input image resolutions to $32\times32$, $64\times64$, and $128\times128$, respectively. The obtained results for each resolution setting are presented in Table \ref{Ablation1}. Based on the acquired results, it is evident that augmenting the dimensions of the input images positively impacts the network's performance. Notably, as the input image transitions from $32\times32$ to $64\times64$, and subsequently to $128\times128$, there is a corresponding enhancement in the PSNR values, with increments of 10\% and 6\%, respectively. Higher-resolution input images exhibit superior network performance albeit at the expense of higher computation resources. Nevertheless, due to the constraints imposed by our computation resources, we were regrettably unable to explore the potential advantages of larger input sizes, such as 256$\times$256 or even 512$\times$512 pixels.

\subsubsection{Hyperrealistic Facial Detail Generator (HFDG).} We subsequently investigate the impact of the HFDG module on the overall performance. We explore three different types of super-resolution networks, namely the GAN-based Real-ESRGAN \cite{wang2021real}, the Transformer-based HAT-B \cite{chen2023activating}, and the Diffusion-based BFRffusion \cite{chen2023towards}. The test results for each distinct architecture of the super-resolution network are summarized in Table \ref{Ablation2}. Based on the obtained results, we can see that HAT-B performs best in the PSNR, SSIM, and CLIPS metrics with an exception in VGGS. Therefore, we carefully considered all factors and ultimately selected HAT-B as the optimal backbone network for the HFDG module in our FRRffusion.

\section{Conclusion}
In this article, we introduced a new computer vision task called Face Retouching Reversal (FRR), which has been increasingly important in our daily lives. Leveraging the powerful commercial APIs, we created a new deepFRR dataset, which became the first FRR dataset. Based on the deepFRR dataset, we proposed an innovative solution by designing the FRRffusion network, which comprises two crucial components: FMAR and HFDG. Extensive experimental results validate the feasibility and efficacy of our approach in addressing the FRR problem. Furthermore, we conducted ablation experiments to gain insights into the parameter selection process for FRRffusion, providing valuable guidance for further exploration of the FRR problem. Notably, since current face retouching APIs work in the local facial regions with skillful algorithms, neither conventional handcrafted image restoration can capture the varying operations of different APIs to perform an effect restoration, nor existing deep methods like Stable Diffusion and GP-UNIT can generate satisfactory results close to the raw image. Promisingly, our FRRffusion digs into the potential of the conditioned diffusion model in a coarse-to-fine manner, providing a new paradigm for handling the FRR problem.

It is worth noting that, as a novel problem and its corresponding solution, this work exhibits several limitations to be resolved in the future. Firstly, the dataset created in this study is relatively limited in terms of the variations in face-retouching effects. To overcome this limitation, we will seek more face-retouching APIs over the Internet for the diversity of the FRR dataset. Secondly, our study casts a new problem where there exists a big gap between the subjective and objective evaluations of the FRR task. How to seamlessly bridge this gap will be a vital research direction. Thirdly, similar to the deep classification and regression models, FRRffusion might be vulnerable to attacks from adversarial examples. Adversarial perturbations could only alter the restored details, implicating an important research field where robustness against adversarial attacks is extremely crucial in such security-sensitive applications. Fourthly, the rapid advent of generative AI models (Midjourney, Sora, etc.) can extend the dataset in scale, which deserves further effort, but might be difficult to be exhausted. Fifthly, we do not provide the performance results on low-resolution input images. This is because the diffusion model is particularly good at learning the data distribution. And our work shows effective restoration for high-resolution (high-dimensional) images. Therefore, we can predict that our FRRffusion will have a better effect on low-resolution input images since low-dimensional distribution is usually more easily learned than high-dimensional ones. Sixthly, we adopted the basic diffusion model (DDPM) to implement our FRRffusion. Considering various variants of DDPM, such as DDIM \cite{song2021denoising}, OMS-DPM \cite{liu2023oms}, and Diff-Pruning \cite{NEURIPS2023_35c1d69d}, our FRRffusion can be greatly boosted by resorting to these cutting-edge architectures.

% ---- Bibliography ----
%
% BibTeX users should specify bibliography style 'splncs04'.
% References will then be sted and formatted in the correct style.
%
\bibliographystyle{splncs04}
\bibliography{main}
\end{document}